\documentclass[10pt,twocolumn,letterpaper]{article}

\usepackage[top=0.75in,bottom=1in,left=0.68in,right=0.68in,columnsep=0.25in]{geometry}
\usepackage{booktabs}
\usepackage{hyperref}
\usepackage{xcolor}
\usepackage{amsmath}
\usepackage{amssymb}
\usepackage{graphicx}
\usepackage{multirow}
\usepackage{times}

\usepackage{titlesec}
\titlespacing*{\section}{0pt}{1.2ex plus 0.4ex minus 0.3ex}{0.6ex plus 0.2ex}
\titlespacing*{\subsection}{0pt}{1.0ex plus 0.3ex minus 0.2ex}{0.5ex plus 0.1ex}
\titlespacing*{\paragraph}{0pt}{0.6ex plus 0.2ex minus 0.2ex}{0.5em}

\hypersetup{
  colorlinks=true,
  linkcolor=blue!60!black,
  citecolor=blue!60!black,
  urlcolor=blue!60!black
}

\title{Storing Less, Finding More: How Novelty Filtering Improves Cross-Modal Retrieval on Edge Cameras\thanks{A video demonstration is included as supplementary material.}}

\author{
  Sherif Abdelwahab \\
  \texttt{sherif@ieee.org}
}

\date{}

\makeatletter
\def\@maketitle{%
  \newpage\null
  \vskip -1.5em
  \begin{center}%
    {\LARGE \@title \par}%
    \vskip 0.8em
    {\large \lineskip .5em \@author \par}%
  \end{center}%
  \vskip 0.8em}
\makeatother

\begin{document}
\maketitle
\vspace{-1em}

\begin{abstract}
Always-on edge cameras generate continuous video streams
where redundant frames degrade cross-modal retrieval by
crowding correct results out of top-$k$ search.  This paper
presents a streaming retrieval architecture: an on-device
$\varepsilon$-net filter retains only semantically novel
frames, building a denoised embedding index; a cross-modal
adapter and cloud re-ranker compensate for the compact
encoder's weak alignment.  A single-pass streaming filter matches or outperforms offline
alternatives (k-means, farthest-point, uniform, random) across
eight vision-language models (8M--632M) on two egocentric
datasets (AEA, EPIC-KITCHENS).  Combined, the
architecture reaches 45.6\% Hit@5 on held-out data using an
8M on-device encoder at an estimated 2.7\,mW.
\end{abstract}

\section{Introduction}

A pair of AR glasses recording at 5\,FPS generates
40\,GB/day~\cite{lee2025aria2}.  A body camera on a 12-hour
shift, a dashcam on a delivery route, a warehouse robot on
patrol: each produces a continuous video stream that someone
will need to search later.  Without on-device filtering, storing or transmitting this
volume is prohibitively expensive, and searching the raw
stream produces degraded results due to redundancy.  Each frame must be kept or discarded the moment
it is captured.

\paragraph{The cross-modal temporal challenge.}
For these devices, retrieval is cross-modal and temporal: the
device stores image embeddings, but the user searches with text
(``when did I leave the kitchen?''), expecting a moment in time.
Large vision-language models bridge the image-text gap but
exceed the power budget of battery-operated devices.  Compact
models fit the power budget but have weaker cross-modal
alignment, making retrieval worse on the small devices that
need it most.

Yet no existing approach is simultaneously streaming,
query-agnostic, and retrieval-improving.  Lowering the frame
rate keeps duplicates and drops unique moments.  Shot-boundary
detectors~\cite{soucek2020transnetv2} need an offline pass.
Summarizers~\cite{narasimhan2021clipit,liang2024keyvideollm}
need the query at selection time.  Training-data
deduplication~\cite{lee2022dedup} and semantic redundancy
removal~\cite{abbas2023semdedup} improve model quality, but no
prior work has shown that removing redundant entries from an
\emph{embedding index} improves retrieval.

\paragraph{Three findings.}
\textbf{(1)~Denoised retrieval:} removing semantically redundant
frames from an embedding index improves cross-modal retrieval,
especially for compact encoders.  Redundant frames crowd correct
results out of top-$k$ nearest-neighbor search; removing them
eliminates this geometric bias.
\textbf{(2)~Streaming suffices:} a single-pass filter, even
with no lookahead, outperforms offline k-means in seven of
eight models and farthest-point sampling in six of eight,
despite seeing each frame only once.
\textbf{(3)~Embeddings-only operation:} because the cross-modal
adapter operates in embedding space, search requires only
compact embeddings (2\,KB per frame), not necessarily raw images
($\sim$150\,KB).  Combined with the filter's $14\times$ frame
reduction, this tightens the bandwidth required for sync,
reduces radio activation time, and extends battery life.

\begin{figure*}[t]
  \centering
  \includegraphics[width=0.85\textwidth]{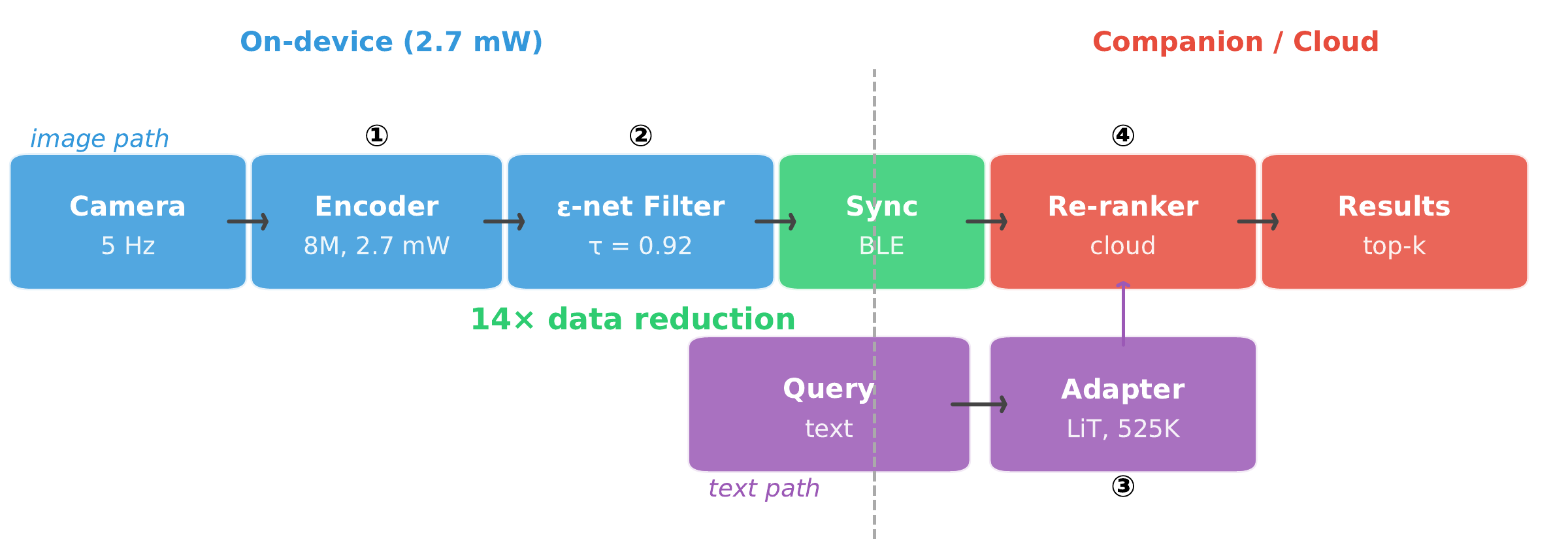}
  \caption{Architecture overview.  On-device (blue): encode and filter frames.
  Query time (purple): text query projected into image space via LiT adapter,
  re-ranked in the cloud.  Circled numbers match stages in the text.}
  \label{fig:system}
\end{figure*}

\paragraph{Architecture.}
These findings make a compact on-device encoder viable and
enable end-to-end retrieval with minimal raw data transmission.
The architecture (Figure~\ref{fig:system}) splits retrieval into
on-device filtering and cloud-side search across four composable,
independently replaceable stages:
\textbf{(1)}~on-device encoder: maps each frame to a compact
embedding (2\,KB), replaceable as better models become
available;
\textbf{(2)}~streaming filter: compares each embedding to
those already retained and discards redundant frames, so
duplicates do not crowd out correct results in top-$k$
search ($14\times$ average reduction on AEA at $\tau = 0.92$
and 5\,FPS; 2.7\,mW total for encoder + filter);
\textbf{(3)}~cross-modal adapter: a small multi-layer perceptron (MLP) trained on
dense egocentric captions projects a larger frozen text
encoder into the compact image space, improving cross-modal
alignment~\cite{zhai2022lit};
\textbf{(4)}~cloud re-ranker: rescores candidates with a
larger model to compensate for the small encoder's residual
alignment gap ($5.7\times$ fewer candidates at equal hit rate).
Only (1) and (2) run on-device.  Because only embeddings
leave the device, no raw visual content is transmitted,
preserving visual privacy by design.  \S\ref{sec:related} surveys related work;
\S\ref{sec:method}--\ref{sec:retrieval} describe the method
and validate across two datasets; \S\ref{sec:pipeline}--\ref{sec:limits} analyze deployment
cost and limitations.

\section{Related Work}
\label{sec:related}

This work draws on frame selection, vision-language retrieval
under edge constraints, and edge-cloud compute partitioning.
None of these areas addresses the combination of streaming,
query-agnostic, and retrieval-improving requirements imposed by
always-on edge devices operating within a $\sim$200\,mW
power budget and constrained wireless
bandwidth~\cite{lee2025aria2}.

\paragraph{Frame selection.}
Shot-boundary detectors~\cite{soucek2020transnetv2,
potapov2014category} and language-guided
summarizers~\cite{narasimhan2021clipit,liang2024keyvideollm}
require an offline pass or the query at selection time; coreset
methods~\cite{gonzalez1985clustering} need all data upfront.
The $\varepsilon$-net filter achieves comparable coverage in a
single streaming pass, as always-on devices cannot buffer or
revisit frames.

\paragraph{Redundancy removal.}
Training-data deduplication~\cite{lee2022dedup} and semantic
redundancy removal~\cite{abbas2023semdedup} improve model
quality.  A key finding here is that removing redundancy from
an inference-time embedding index improves \emph{retrieval},
through a different mechanism: geometric bias in cross-modal
search (\S\ref{sec:retrieval}).

\paragraph{Compact encoders and alignment.}
Distilled encoders (TinyCLIP~\cite{wu2023tinyclip},
MobileCLIP~\cite{vasu2024mobileclip}) trade alignment for
size; Locked-image Tuning (LiT)~\cite{zhai2022lit} recovers
alignment by training only
the text encoder; ClipBERT~\cite{lei2021clipbert} validates sparse
sampling.  This architecture uses a distilled encoder, a LiT-style
adapter, and sparse frame retention, and adds index-side
semantic redundancy filtering that improves retrieval for any encoder.

\paragraph{Edge-cloud placement and egocentric benchmarks.}
Compute placement is the central tradeoff in edge
AI~\cite{chen2025edgeai,wang2025ondevice,lee2025aria2}: moving
primitives on-device does not always save power.  This
architecture keeps only the filter and encoder on-device.  Ego4D~\cite{grauman2022ego4d} and
EgoVLPv2~\cite{pramanick2023egovlpv2} provide egocentric
benchmarks focused on better models; this work is orthogonal,
improving retrieval through index-side redundancy filtering under an edge
constraint these benchmarks do not impose.  Video moment
retrieval methods (e.g.\ temporal grounding, highlight
detection) assume access to the full video at query time;
this architecture must select frames before any query is
known.

\section{Method}
\label{sec:method}

Let $\{f_1, \ldots, f_N\}$ be frames from an egocentric camera. An
embedding model $\phi$ maps each frame to a unit vector
$\mathbf{v}_i = \phi(f_i) \in \mathbb{R}^d$. A temporal memory
$\mathcal{M}$ stores embeddings of retained frames. Frame $f_i$ is
kept if:
\begin{equation}
\max_{\mathbf{m} \in \mathcal{M}} \; \mathbf{v}_i^\top \mathbf{m} < \tau
\label{eq:selection}
\end{equation}
Otherwise the frame is skipped. When a frame is kept, its embedding
$\mathbf{v}_i$ is added to $\mathcal{M}$. The first frame is always
kept.

The threshold $\tau$ is the sole parameter.  Processing each frame
requires one forward pass through the embedding model and one
image-to-image nearest-neighbor lookup in $\mathcal{M}$ (no text is
involved at selection time).  The algorithm is single-pass with no
lookahead.

By construction, the retained set $\mathcal{K}$ forms a maximal
$\varepsilon$-net with $\varepsilon = 1 - \tau$: every
discarded frame lies within cosine distance $\varepsilon$ of
a retained keyframe (coverage), and no two keyframes are
closer than $\varepsilon$ (separation).  A different frame
ordering could yield a different net, but the coverage
guarantee holds regardless.

\paragraph{Why filtering improves retrieval.}
In top-$k$ retrieval, an event with many near-duplicate frames
can fill most result slots even when a correct event is closer
to the query.  The $\varepsilon$-net equalizes frame counts
across events, so ranking shifts from cluster size to semantic
distance.  The effect is cross-modal, unsupervised, and
model-agnostic (\S\ref{sec:retrieval}).

\paragraph{Text adaptation via LiT.}
At query time, a Locked-image Tuning
(LiT)~\cite{zhai2022lit} adapter aligns the text query to the
frozen image embedding space.  The adapter is a 525K-parameter
residual MLP applied after a frozen CLIP text encoder, trained
on dense egocentric captions paired with image embeddings.  It runs on a companion device (e.g.\ a paired phone) or
cloud endpoint, not on the camera.

\section{Experimental Setup}

The evaluation tests whether the denoising effect holds across
models, datasets, and selection algorithms.  It uses 19
recordings from Aria Everyday Activities
(AEA)~\cite{lv2024aria} (85,002 frames, 553 events, five
locations) and 7 EPIC-KITCHENS
videos~\cite{damen2022rescaling} (median event 1.7\,s,
\S\ref{sec:tau}).

\paragraph{Two-step protocol.}
Selection and retrieval use separate models.
CLIP ViT-B/32~\cite{radford2021learning}, the smallest
contrastive model with sufficient alignment for the
$\varepsilon$-net filter and therefore the closest proxy to
target edge device classes, selects keyframes ($\tau = 0.92$).
Fixing the selector isolates the denoising effect from
encoder quality: each of eight independently trained models
(Table~\ref{tab:models}, 8M--632M) then \emph{re-embeds the
selected frames in its own space} and retrieves against text
queries.  No embeddings cross model boundaries.  That all
eight benefit, including models much larger than the selector,
confirms the effect is geometric rather than an artifact of
B/32's space.  In deployment, a single on-device encoder
would both select and embed.

\paragraph{Retrieval metric.}
For each annotated event, the model embeds the event's text
description, retrieves top-5 frames by cosine similarity, and
a retrieval counts as correct (a \emph{hit}) if any returned
frame falls within $\pm 0.5$\,s of the annotated timestamp.
All reported percentages are Hit@5: the fraction of events
with at least one correct frame in the top-5 results.

\paragraph{Selection strategies.}
Six strategies are compared at matched frame counts:
keeping all frames (\emph{Full}), retaining only novel frames
via the streaming $\varepsilon$-net (\emph{Novelty}), selecting
cluster centroids via offline k-means (\emph{K-Means}),
greedy farthest-point sampling~\cite{gonzalez1985clustering}
(\emph{FP}), taking every $n$-th frame (\emph{Uniform}), and
picking frames at random (\emph{Random}).  Matching the frame
count ensures that differences reflect selection quality, not
compression ratio.

\begin{table}[htbp]
\caption{Evaluation models (8M--632M params, four training
paradigms). Each evaluates retrieval independently.}
\label{tab:models}
\centering
\footnotesize
\begin{tabular}{@{}lrlr@{}}
\toprule
Model & Params & Training & Dim \\
\midrule
TinyCLIP~\cite{wu2023tinyclip} & 8M & CLIP distillation & 512 \\
MobileCLIP-S0~\cite{vasu2024mobileclip} & 11M & CLIP distillation & 512 \\
CLIP ViT-B/32$^\dagger$~\cite{radford2021learning} & 88M & Contrastive & 512 \\
CLIP ViT-L/14~\cite{radford2021learning} & 304M & Contrastive & 768 \\
SigLIP ViT-B/16~\cite{zhai2023sigmoid} & 86M & Sigmoid & 768 \\
SigLIP\,2 SO400M~\cite{tschannen2025siglip2} & 400M & Sigmoid+caption & 1152 \\
ImageBind~\cite{girdhar2023imagebind} & 632M & 6-modality & 1024 \\
\midrule
TinyCLIP + LiT & 8M+0.5M & + adapter (proposed) & 512 \\
\bottomrule
\end{tabular}
\end{table}

\begin{figure*}[t]
\centering
\includegraphics[width=0.85\textwidth]{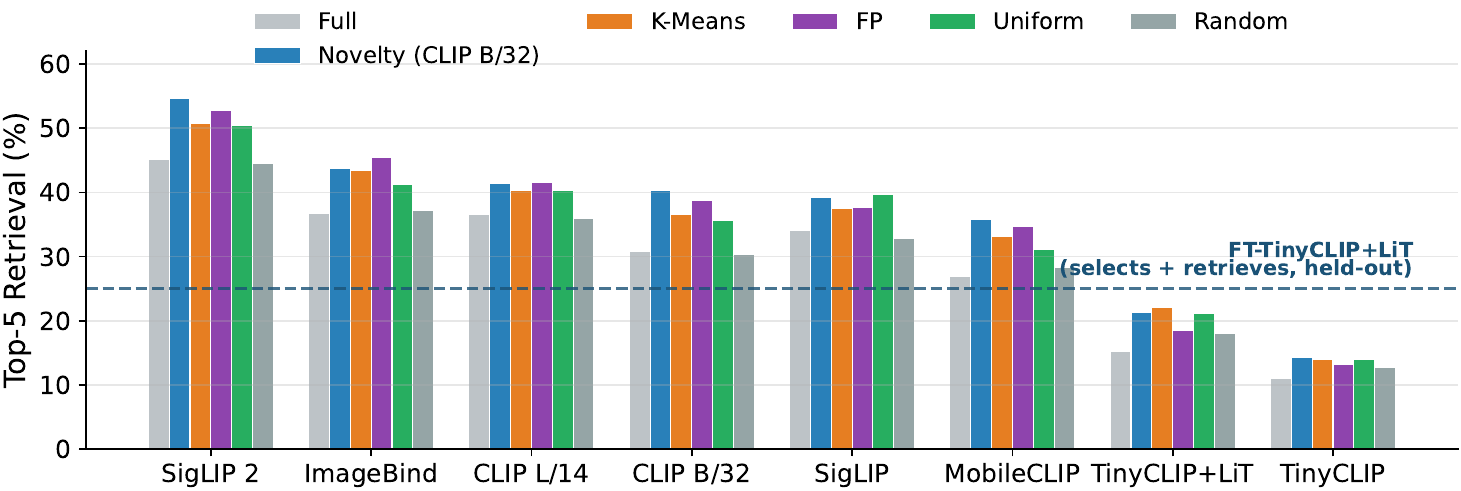}
\caption{Top-5 retrieval on 19 AEA sequences (553 events).
CLIP\,B/32 selects which frames to keep; each model on the
x-axis re-embeds those frames in its own space and retrieves
independently.  Novelty outperforms all offline baselines in five of eight models.  Dashed line: FT-TinyCLIP+LiT on held-out data
(25.0\%).}
\label{fig:retrieval}
\end{figure*}

\section{Results}

\subsection{Does filtering improve retrieval?}
\label{sec:retrieval}

Table~\ref{tab:main} and Figure~\ref{fig:retrieval} show that
searching the filtered embedding index retrieves more events than
searching the full frame set, across all eight models (six
statistically significant, $p < 0.05$).  The filtered index also
outperforms offline k-means (seven of eight models),
farthest-point sampling (six of eight), and uniform sampling
(seven of eight).  The retrieval improvement from filtering (Novelty minus Full,
in absolute percentage points) is largest for compact models
with weaker cross-modal alignment: CLIP\,B/32 (+9.6\,pp),
MobileCLIP (+8.9\,pp), TinyCLIP (+3.3\,pp).  Filtering helps
these models most, but their absolute top-5 retrieval remains
low (TinyCLIP: 14.3\%); \S5.2 addresses this.  Two models
(CLIP L/14, SigLIP) do not reach significance at $p < 0.05$;
the improvement is directionally consistent but the sample
size (19 sequences) limits statistical power.  TinyCLIP+LiT
is the only model where k-means (22.1) edges out novelty
(21.3); the 0.8\,pp gap is within noise, and k-means requires
an offline pass over all frames while the streaming filter
operates in a single pass.  Farthest-point wins on the two
largest models (ImageBind, CLIP\,L/14) where alignment is
already strong; novelty wins on the compact models where
alignment is weak, which is the on-device deployment target.

\begin{table}[htbp]
  \caption{Filtering improves retrieval across all eight models.
  CLIP\,B/32 selects frames; each model re-embeds and retrieves
  independently.  Top-5 (\%), AEA, 553 events, matched frame
  count.}
  \label{tab:main}
  \centering
  \footnotesize
  \setlength{\tabcolsep}{3pt}
  \begin{tabular}{@{}lrrrrrrc@{}}
  \toprule
  Model & Full & Nov & KM & FP & Uni & Rnd & $p$ \\
  \midrule
  SigLIP\,2       & 45.2 & \textbf{54.6} & 50.8 & 52.8 & 50.5 & 44.6 & .003 \\
  ImageBind        & 36.7 & 43.8 & 43.4 & \textbf{45.4} & 41.2 & 37.2 & .007 \\
  CLIP L/14        & 36.5 & 41.4 & 40.3 & \textbf{41.6} & 40.3 & 35.9 & .12 \\
  CLIP B/32$^\dagger$ & 30.7 & \textbf{40.3} & 36.5 & 38.7 & 35.6 & 30.4 & $<$\!.001 \\
  SigLIP           & 34.0 & 39.2 & 37.4 & 37.6 & \textbf{39.6} & 32.8 & .27 \\
  MobileCLIP       & 26.9 & \textbf{35.8} & 33.1 & 34.7 & 31.1 & 28.3 & $<$\!.001 \\
  TinyCLIP+LiT     & 15.2 & 21.3 & \textbf{22.1} & 18.4 & 21.2 & 18.0 & .002 \\
  TinyCLIP         & 11.0 & \textbf{14.3} & 13.9 & 13.2 & 13.9 & 12.8 & .021 \\
  \bottomrule
  \end{tabular}\\[2pt]
  {\scriptsize All values are percentages.  Nov: Novelty ($\varepsilon$-net).  KM: K-Means.  FP: Farthest-Point.  Uni: Uniform.  Rnd: Random.  $p$: two-sided sign test (novelty vs.\ full).  $^\dagger$Selector model.  }
  \end{table}

\paragraph{Robustness.}
Bootstrap resampling (1000 trials over sequences) confirms
stability: the novelty gain stays positive in every trial,
with a worst case of $+3.3$\,pp and median $+6.3$ to
$+9.7$\,pp across four models.  Cross-dataset generalization
(EPIC-KITCHENS) is tested in \S\ref{sec:tau}.

\subsection{Can a LiT adapter close the alignment gap?}

Filtering alone leaves TinyCLIP at 14.3\% top-5 retrieval.
Table~\ref{tab:4way} shows a second, independent lever:
the LiT adapter (\S\ref{sec:method}) improves cross-modal
alignment ($+4.2$\,pp) without changing the on-device encoder.
The two gains multiply (filtering: $\times 1.30$, adapter:
$\times 1.38$) because they act on different sides of
retrieval: filtering cleans the index, the adapter improves
the query.  Combined, top-5 rises from 11.0\% to 21.3\%.

\begin{table}[htbp]
  \caption{TinyCLIP 4-way: filtering ($\times 1.30$) and
  text adaptation ($\times 1.38$) multiply independently.}
  \label{tab:4way}
  \centering
  \footnotesize
  \begin{tabular}{@{}llcc@{}}
  \toprule
  Image index & Text encoder & Top-5 & $\Delta$ \\
  \midrule
  All frames & CLIP B/32 text & 11.0\% & --- \\
  $\varepsilon$-net & CLIP B/32 text & 14.3\% & +3.3 \\
  All frames & + LiT adapter & 15.2\% & +4.2 \\
  $\varepsilon$-net & + LiT adapter & \textbf{21.3\%} & +10.3 \\
  \bottomrule
  \end{tabular}
  \end{table}
  
\subsection{Does domain fine-tuning help?}
\label{sec:finetune}

The off-the-shelf adapter reaches 21.3\% on all 19
sequences.  To push further, TinyCLIP's image encoder and
the LiT adapter are fine-tuned jointly (both trained
simultaneously).  Each retained frame is paired with a dense
caption from a vision-language model (Gemini Flash); the
training target is the SigLIP\,2 text embedding of that
caption, chosen as the strongest-aligned model
(Table~\ref{tab:main}, 54.6\%).  An 80/20 split (15 AEA + 5 EPIC training, 6 AEA + 2 EPIC
held out; 10,911 pairs) measures generalization.

\begin{table}[htbp]
\caption{Joint fine-tuning (top-5, $\varepsilon$-net index).}
\label{tab:finetune}
\centering
\footnotesize
\begin{tabular}{@{}llcc@{}}
\toprule
Split & Configuration & Top-5 & $\Delta$ \\
\midrule
\multirow{2}{*}{AEA test (unseen)} & TinyCLIP & 16.9\% & --- \\
 & FT-TinyCLIP+LiT & \textbf{25.0\%} & +8.1 \\
\midrule
\multirow{2}{*}{AEA train (seen)} & TinyCLIP & 13.5\% & --- \\
 & FT-TinyCLIP+LiT & 48.3\% & +34.7 \\
\midrule
\multirow{2}{*}{EPIC test (unseen)} & TinyCLIP & 1.6\% & --- \\
 & FT-TinyCLIP+LiT & \textbf{6.0\%} & +4.3 \\
\bottomrule
\end{tabular}
\end{table}

On held-out AEA, retrieval improves from 16.9\% to 25.0\%
(+48\% relative); on held-out EPIC, from 1.6\% to 6.0\%.
Cross-validation confirms generalization: training on 10 AEA
sequences and evaluating on the other 9 yields 28.3\%.  The
23-point train/test gap (48.3\% vs.\ 25.0\%) reflects the
small training set, but the held-out gain is consistent
across all three generalization tests.

\subsection{Does the effect generalize across datasets?}
\label{sec:tau}

All results so far use AEA, where events last a median of
6.0\,s.  At $\tau = 0.92$ the filter retains enough frames to
cover these long events.  EPIC-KITCHENS events are much
shorter (median 1.7\,s): at the same $\tau$, the filter
discards too aggressively and drops brief actions.
Table~\ref{tab:tau} sweeps $\tau$ on 7 EPIC videos.  Below
0.94, novelty wins only 3--4 of 7 videos; at $\tau \geq 0.94$
the filter retains enough frames for short events and
novelty wins 6 of 7.  The absolute numbers are low across
all methods because EPIC events average 1.7\,s (roughly 8
frames at 5\,FPS), making each event a small retrieval
target; the full-set baseline is itself only 5.8\%.  The
threshold is dataset-dependent, but once matched to event
duration the denoising effect generalizes.

\paragraph{Dataset properties shape conclusions.}
AEA's near-complete annotation coverage (97.3\%) and long
events (median 6.0\,s) flatten the difference between
selection methods: even uniform sampling performs well because
most frames fall within some annotated event.  EPIC's sparser
coverage (82\%) and shorter events (median 1.7\,s)
differentiate selection methods more sharply.  The choice of
evaluation dataset can determine which method appears to win.

\begin{table}[htbp]
\caption{EPIC-KITCHENS threshold sweep (CLIP\,B/32, top-5).}
\label{tab:tau}
\centering
\footnotesize
\begin{tabular}{@{}ccrrrr@{}}
\toprule
$\tau$ & Comp & Novelty & Uniform & Full & Nov$>$Full \\
\midrule
0.90 & 35$\times$ & 5.6\% & 5.4\% & 5.8\% & 3/7 \\
0.92 & 16$\times$ & 6.0\% & 5.6\% & 5.8\% & 4/7 \\
0.94 &  7$\times$ & 6.6\% & 6.4\% & 5.8\% & 6/7 \\
0.95 &  4$\times$ & \textbf{6.7\%} & 6.4\% & 5.8\% & 6/7 \\
0.96 &  3$\times$ & \textbf{6.8\%} & 6.0\% & 5.8\% & 6/7 \\
\bottomrule
\end{tabular}\\[2pt]
{\scriptsize Full-set baseline: 5.8\%.  Nov$>$Full: per-video wins out of 7.}
\end{table}

\subsection{What does it cost to deploy?}

An always-on edge device (e.g.\ a wearable like
Aria Gen~2) budgets $\sim$200\,mW for always-on
compute~\cite{lee2025aria2}.  TinyCLIP (INT8, ONNX) profiled
on a Hexagon v69 NPU ($\sim$2.5\,TOPS) via Qualcomm AI Hub
completes inference in 2.16\,ms per frame (8.1\,MB model,
100\% NPU utilization), consuming $\sim$1\% of available NPU
cycles at 5\,FPS.  A conservative back-of-the-envelope power estimate based
on the NPU datasheet gives $\sim$2.7\,mW per frame; actual
power depends on the SoC and is not measured here, but the
cycle count confirms the workload is small relative to what
sub-5\,TOPS edge NPUs provide.

Because the architecture operates on embeddings, not raw
images, only 2\,KB per retained frame needs to leave the
device (vs.\ $\sim$150\,KB for a raw frame).  The
$14\times$ filtering at 5\,FPS reduces the outgoing data to
an estimated 43\,KB/min (39\,MB/day), making BLE sync
plausible at $\sim$21\% radio duty cycle.\footnote{43\,KB/min
$\div$ $\sim$200\,KB/min effective BLE\,4.0 throughput
(20-byte payloads, connection overhead) $\approx$ 21\%.
BLE\,5.0 would reduce this further.}  Without filtering, the raw
stream is $\sim$45\,MB/min, far beyond BLE capacity.  These
are conservative estimates; actual throughput depends on
radio conditions and BLE stack overhead.  This embeddings-only
mode also preserves visual privacy: no raw images leave the
device, only 512-dimensional vectors from which reconstructing the
original frame is impractical.

\subsection{How does the full pipeline perform?}
\label{sec:pipeline}

\begin{figure*}[t]
\centering
\includegraphics[width=\textwidth]{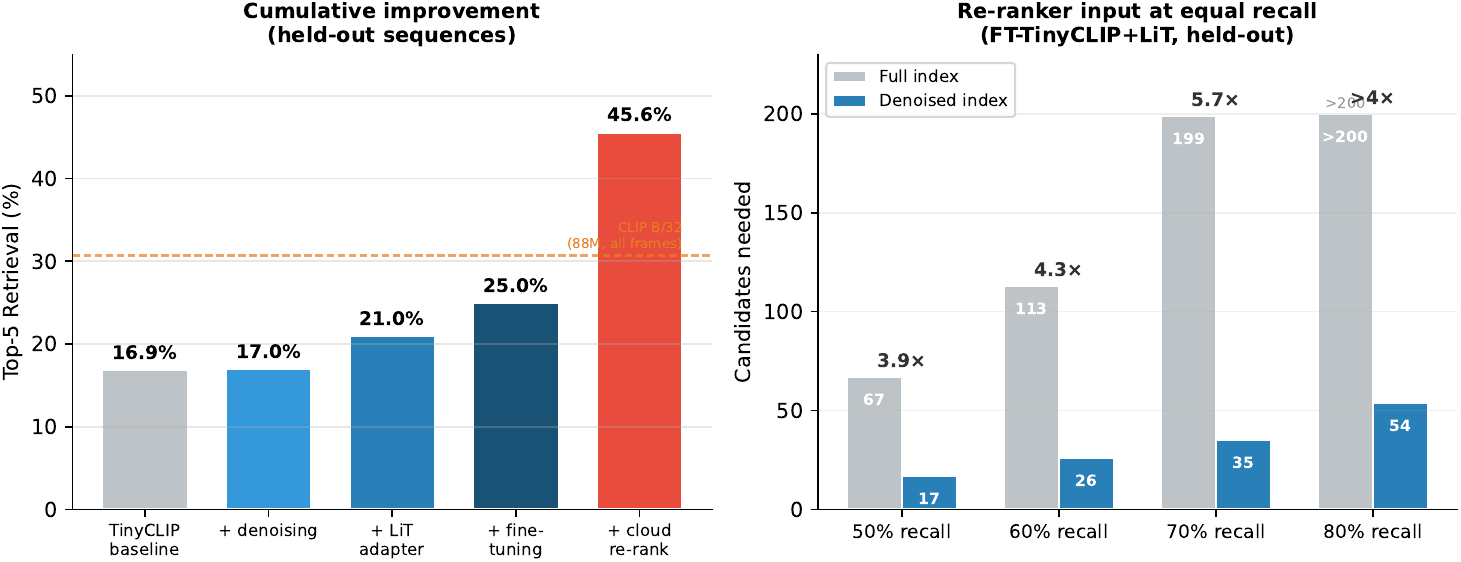}
\caption{Left: cumulative improvement on held-out data (dashed
line: CLIP\,B/32 on all frames).  Right: re-ranker candidates
needed at equal hit rate ($5.7\times$ fewer at 70\%).}
\label{fig:results}
\end{figure*}

After filtering and fine-tuning, the on-device encoder
reaches 25.0\% on held-out data.  For a practical system
this is still low, but the on-device encoder cannot grow
larger without exceeding the power budget.  Because the
architecture separates device-time encoding from query-time
search, a larger model can rescore the shortlist without
touching the edge device.  SigLIP\,2 (400M), the
strongest-aligned model in our evaluation
(Table~\ref{tab:main}), rescores the top-50 candidates and
pushes the final top-5 to 45.6\%, exceeding CLIP\,B/32 on
all frames (30.7\%) despite an 8M on-device encoder.
Figure~\ref{fig:results} (left) shows each stage composed
cumulatively on held-out data.

Figure~\ref{fig:results} (right) shows why the filtered index
is a better input to the re-ranker: at equal hit rate, the
denoised index requires $5.7\times$ fewer candidates.  The
filtering gain also grows with retrieval depth: at top-50
(Hit@50), the denoised index finds 77.9\% of events vs.\
33.9\% for the unfiltered index; in retrieval-system terms,
the re-ranker receives 2.3$\times$ more correct candidates
from the denoised first stage.  The on-device filter is
fixed; better
re-rankers or larger $k$ improve the architecture without
changing the device.

\section{Limitations and Future Work}
\label{sec:limits}

\paragraph{Evaluation scope.}
Both datasets (AEA, EPIC-KITCHENS) are kitchen-heavy.
Validation on outdoor, industrial, or multi-room environments
would test whether the denoising effect holds when visual
diversity increases.

\paragraph{Fixed threshold and compression drift.}
$\tau$ must be matched to event duration (\S\ref{sec:tau});
a single value does not work across datasets.  Furthermore,
as the keyframe buffer grows, the probability of a new frame
passing the novelty check decreases monotonically, so the
effective compression ratio increases over time.  On a
36-minute EPIC video (P22\_07), this produced the worst
novelty-vs-full degradation in the evaluation.  Adaptive
algorithms that adjust $\tau$ online are future work.

\paragraph{Overfitting gap.}
The 23-point train/test gap (48.3\% vs.\ 25.0\%) reflects a
training set of 10,911 pairs from 20 sequences.  More
training data from diverse environments should narrow this gap.

\paragraph{Tolerance window.}
The $\pm 0.5$\,s tolerance is not swept; tighter tolerances
would test whether the denoising gain depends on temporal
margin.

\paragraph{Sensor and memory.}
The filter uses only the RGB stream; multi-sensor gating
(IMU, audio) could suppress embedding computation during
stationary periods.  The filter does not detect temporal
re-entry (returning to a previously visited scene).

\section{Conclusion}

Redundant frames are not just a storage problem; they are a
retrieval problem.  Removing them from an embedding index
eliminates a geometric bias in nearest-neighbor search that
degrades cross-modal retrieval, especially for the compact
encoders that edge devices require.

This paper showed that a single-pass streaming filter, the
simplest possible selection algorithm, produces a better
embedding index than offline alternatives that see all frames
before selecting.  Combined with a cross-modal adapter and
cloud re-ranker, the architecture reaches 45.6\% Hit@5 on
held-out data (77.9\% Hit@50) using an 8M-parameter on-device
encoder at an estimated 2.7\,mW.

All four stages are independently replaceable.  Better
encoders, adapters, or re-rankers improve the architecture
without redesigning the device.  The on-device filter is
fixed at deployment; everything else can be upgraded over the
air.

{\small
\bibliographystyle{plain}
\bibliography{references}
}

\end{document}